\documentclass[letterpaper]{article} 
 \newlength\titlebox 
 \twocolumn 
\usepackage{times} 
\usepackage{helvet}  
\usepackage{courier} 
\usepackage[hyphens]{url} 
\usepackage{graphicx}
\usepackage{caption} 
\usepackage{algorithm}
\usepackage{algorithmic}

\usepackage{newfloat}
\usepackage{listings}

\usepackage{newfloat}
\usepackage[flushleft]{threeparttable}
\usepackage{color}
\usepackage{subfig}
\usepackage{soul}
\usepackage{booktabs}
\usepackage{array, tabularx, caption, boldline}
\usepackage{graphicx}
\usepackage{cellspace}
\usepackage{amsthm}
\usepackage{listings}
\usepackage{tabularx} 
\usepackage{method}
\usepackage{algorithm}
\usepackage{amsmath}
\usepackage{amssymb}
\usepackage{multirow}
\usepackage[table]{xcolor}
\usepackage[export]{adjustbox}
\usepackage{makecell}

\title{Outlier Ranking in Large-Scale Public Health Streams}
\author{Ananya Joshi\thanks{Corresponding Author Email: aajoshi@andrew.cmu.edu}, Tina Townes, Nolan Gormley, Luke Neureiter, Roni Rosenfeld, Bryan Wilder}

\date{%
    Carnegie Mellon University\\%
    5000 Forbes Rd.\\%
    Pittsburgh, Pennsylvania, 15213, USA\\%
}

\begin{document}
\maketitle
\begin{abstract}
Disease control experts inspect public health data streams daily for outliers worth investigating, like those corresponding to data quality issues or disease outbreaks. However, they can only examine a few of the thousands of maximally-tied outliers returned by univariate outlier detection methods applied to large-scale public health data streams. To help experts distinguish the most important outliers from these thousands of tied outliers, we propose a new task for algorithms to \textbf{rank} the outputs of any univariate method applied to each of many streams. Our novel algorithm for this task, which leverages hierarchical networks and extreme value analysis, performed the best across traditional outlier detection metrics in a human-expert evaluation using public health data streams. Most importantly, experts have used our open-source Python implementation since April 2023 and report identifying outliers worth investigating \textbf{9.1x} faster than their prior baseline. Other organizations can readily adapt this implementation to create rankings from the outputs of their tailored univariate methods across large-scale streams. 
\end{abstract}

\section{Motivation}
Disease control experts must analyze large volumes of time-series population-level public health data (streams) for data delays, errors, or changes in disease dynamics to mitigate disease spread \cite{WHO, UN, CDC_surv}. Because these streams are noisy, nonstationary, and subject to different dynamics, experts can most clearly detect these phenomena when they manifest as specific types of point outliers (e.g., large spikes) in individual data streams \footnote{The joint relationships between streams are generally unstable due to the statistical properties of public health data streams.}. Thus, experts have relied on tailored \textit{univariate} methods \cite{blazquez2021review} applied independently per stream to prioritize data points (outliers) corresponding to important phenomena. 

\begin{figure}
  \centering
\includegraphics[width=7.5cm]{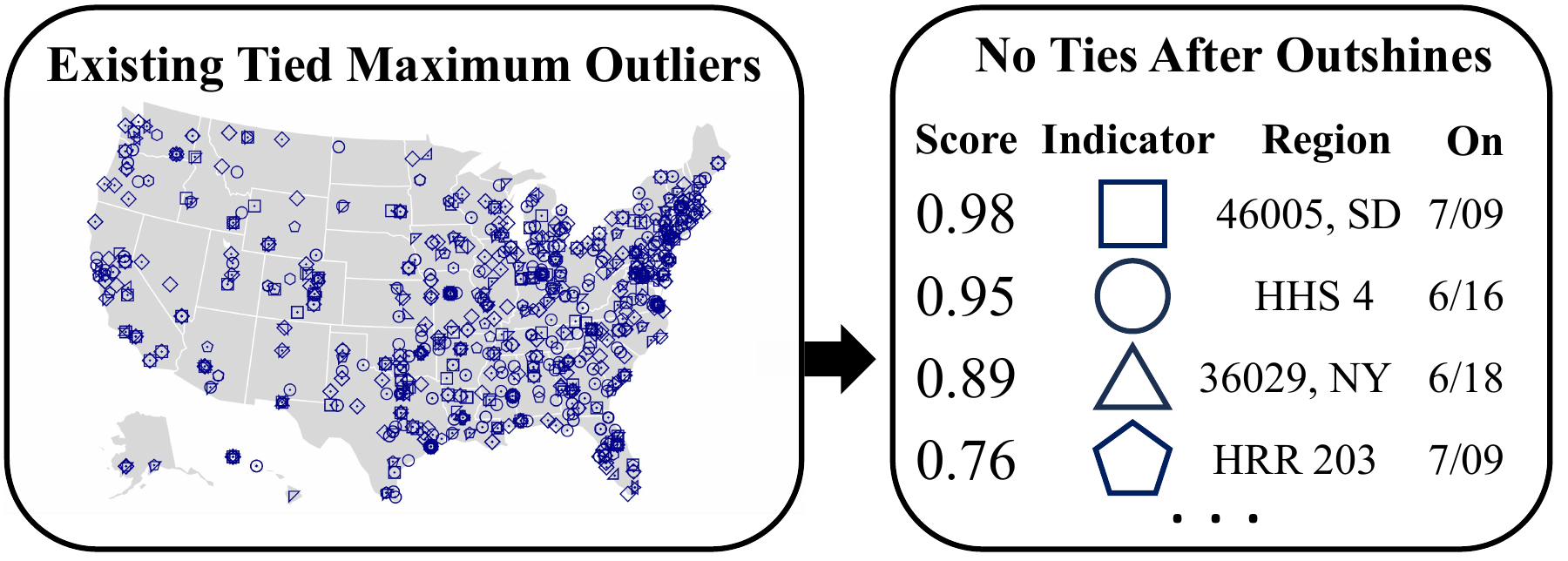}
  \caption{Existing methods identify 14-20k tied maximum-priority outliers (a subset shown here) from 3-4m points. Outshines ranking enables experts to prioritize outliers.}
  \label{fig:Outshines}
\end{figure}

The rapid expansion of public health data since the start of the COVID-19 pandemic \cite{whostrat} exposed a fundamental flaw in this approach. When public health organizations only curated a few thousand weekly-updated data streams, the number of maximum-priority outliers identified this way was relatively small, and experts could examine all outliers manually. However, with modern volumes of public health data containing hundreds of thousands of data streams updated daily, this approach returns an overwhelming number of tied maximum-priority outliers (e.g., 14-20k, as shown in Fig. \ref{fig:Outshines}), of which many are false positives.

To address this challenge, we collaborated with the Delphi Group at Carnegie Mellon University. Delphi has curated public health data for nearly a decade and maintains an API used by various public health organizations \footnote{These organizations are generally in the United States (U.S.) and include national and local organizations.}. In the past three years, Delphi's daily data intake increased over \textbf{1000x} to 3-4 million data points across 200-300k data streams due to the availability of public health data indicators, including official indicators, like cases or hospitalizations, and auxiliary indicators, like insurance claims and cellphone-based mobility \cite{whopret, kraemer2021data}. These indicators are also available at a higher resolution (e.g., from $\{$U.S. state x weekly$\}$ to $\{$U.S. county x daily$\}$), and with historical data revisions \cite{reinhart2021open}, which may contain new outliers. Existing univariate methods return false positives in this data for three reasons:
\begin{enumerate}
\item Because these methods operate independently per stream, outputs should not be compared across streams. More streams means more potential false positives because each stream has a different outlier threshold.
\item These methods may not account for temporal recency and thus may not be able to appropriately rank outliers in data streams that change rapidly (nonstationary). 
\item Nonstationarity and historical data availability limit the granularity of the output outlier scores (used for ranking).  
\end{enumerate}

Motivated by this setting, we introduce a new task where the goal is to rank the overall highest-priority outliers across all data streams, thus limiting false positives and prioritizing expert time: \textbf{multi-stream outlier ranking}. Algorithms for this task act as a post-processing calibration step by taking values produced by any univariate outlier detection method applied independently to each of a large number of data streams as input and ranking them by leveraging the historical behavior of the underlying univariate method per stream. 
Our novel algorithm for this task, Outshines (\textbf{Out}lier\textbf{s} in \textbf{Hi}erarchical \textbf{Ne}twork\textbf{s}), uses hierarchical relationships between the streams to prioritize outliers that are extreme relative to the recent historical behavior of the univariate method across all data streams. Because Outshines works with any univariate outlier detection method and on different types of indicators, it can adapt to the increasing and changing volumes of large-scale public health data.

Our results from expert evaluations using real public health data streams showed that Outshines performs the best across traditional outlier detection metrics in our comparisons. As a result, Delphi has deployed Outshines, and experts have used it as part of their daily data analysis process since April 2023. These experts report identifying outliers worth investigating 9.1x faster with Outshines. 

The Outshines implementation and evaluation materials are open-source at: \url{https://github.com/Ananya-Joshi/Outshines-Documentation}. While we focus on public health, this task and algorithm may also be relevant in other domains with similar problem formulations (e.g., economics, climate science, manufacturing, fraud detection, and computer systems). 

\section{Problem Background \& Formulation}
\begin{figure}
  \centering
 \includegraphics[width=7cm]{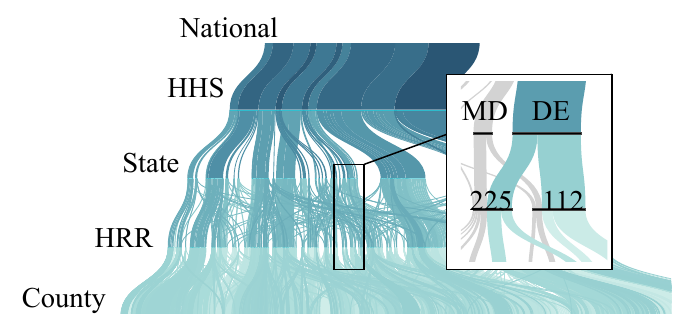}
  \caption{The geospatial hierarchy for public health streams covers 4270 regions. HRRs may serve multiple states like HRR 112 and 225 serve both D.E. and M.D. residents.}
  \label{fig:hiernetemp}
\end{figure}

Each day, $T$, Delphi receives data points $d_{i,r}(t)$, where indicator $i$ is in the set of curated indicators $\mathcal{I}$, $r$ is a region in $\mathcal{R}$, where $\mathcal{R}$ contains all the regions in Fig. \ref{fig:hiernetemp} from different tiers (e.g. county, state or national), and $t$ is a historical day $0 \leq t \leq T$. This geospatial-temporal data forms data streams, where each stream (identified by $i, r$) consists of $d_{i,r}(t) \,\forall\, t \in [0, T]$. The number of data streams is large ($\lvert\mathcal{I}\rvert\times\lvert\mathcal{R} \rvert$) and far exceeds the history ($T$) available per stream. These streams are then optionally processed with a Gaussian smoother and a Poisson regression model to remove weekday effects \cite{reinhart2021open}. Because data providers censor data to prevent reverse identification and may report data with missing values, higher-tiered regions (e.g., states) may include data from more people than expected by combining information from county-level sub-regions and must be individually analyzed.    

\subsubsection{Geospatial Hierarchies}
Regions in $\mathcal{R}$ form a hierarchy (see Fig. \ref{fig:hiernetemp}) which captures geospatial and epidemiological relationships. Notably, hospital referral regions (HRR) share a hospital system \cite{hrr}, and HHS groups contain nearby states \cite{hhs}. Experts use these hierarchical relationships (e.g., parent, \textbf{sibling} (which share a parent), and child streams) for outlier ranking. For example, an expert may rank outliers by comparing them to respective points in sibling streams. 

\begin{figure*}
  \centering
  \includegraphics[width=\textwidth]{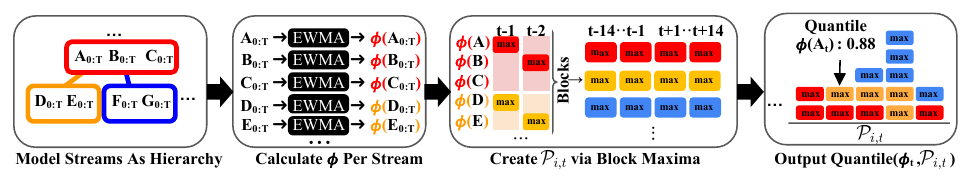}
  \caption{Outshines models the hierarchical relationships between streams, applies a univariate method per stream to calculate $\phi$s, creates $\mathcal{P}_{i, t}$ via block maxima per indicator $i$ per day $t$ across sibling streams, and finally, ranks $\phi_t$ by its quantile in $\mathcal{P}_{i,t}$.}
  \label{fig:Outshinesmet}
\end{figure*}

\paragraph{Problem Formulation:} Our goal is to increase the rate at which experts can identify data points that they deem worth investigating (\textbf{events}) in large-scale data streams. Tailored univariate point outlier detection methods can identify these events, but not at large scales. So that we can use univariate methods scale for large-scale (many) streams, we formalize the multi-stream outlier ranking task, where an algorithm inputs values from univariate methods and outputs rankable scores $y(d)$ over all newly received data $\mathcal{D}$ per day. This ranking can better prioritize expert attention. We assess candidate algorithms using a range of traditional outlier classification and ranking metrics comparing $y$ and expert-provided outlier scores $c(d) \,\forall \,d \in \mathcal{D}$. The implemented algorithm must also meet Delphi's deployment constraint; it must process the complete data volume (including handling changing, missing, and revised data) in less than one day to stay current with daily updated data. 

\section{Related Works}
To our knowledge, previous work has never considered multi-stream outlier ranking. While there are many uni- and multi-variate outlier detection methods \cite{blazquez2021review}, neither is adequate for this problem. 
\textbf{Univariate} outlier detection methods identify outlier data points in individual streams as needed, \cite {rabanser2022intrinsic, hundman2018detecting, blazquez2021review}, can operate over data streams with different properties (e.g. scale\footnote{E.g., the raw COVID case count in a rural county (0-10) will be much differently scaled than that in a city (0-1000)}, noise, and outlier patterns) \cite{cook2019anomaly}, and are fully parallelizable over large sets of streams. They are particularly relevant because public health data streams exhibit strong temporal dependencies. However, these methods currently return too many false positives (Fig. 1). 

On the other hand, while \textbf{multivariate} methods can identify outliers given interconnected factors, they are limited in several fundamental ways. Some methods rely on dimension reduction or sampling \cite{jin2017anomaly, zan2023spatial}, which means that not all data is analyzed. Otherwise, they rely on the correlation structures between the streams, but these are known to change rapidly in noisy and nonstationary data (like public health data) \cite{jadidi2023correlation, gottwalt2019corrcorr}. Multivariate methods also generally require much more computational power than univariate methods (e.g., via large matrix inversions). Unlike the public health setting, these approaches are more relevant in settings where data streams contain similar measurements uniformly impacted by global events. Still, even public health-specific approaches, like outbreak detection \cite{wong2005s, kulldorff2005space, buckeridge2005algorithms}, are limited by the previously described challenges.  

\paragraph{Outlier Ranking Methods:} Outlier ranking methods are overlooked in the existing literature. We identified that outlier methods create ranks from the quantiles of new data points with respect to an empirical reference distribution, $\mathcal{P}$. This distribution, $\mathcal{P}$, uses historical data from a single stream or a small set of similar streams. As the size of $\mathcal{P}$ increases, so does the resolution of quantiles possible that determine empirical scores used for outlier ranking. For example, if $\mathcal{P}$ only has 10 values, the quantiles would correspond to deciles. The more quantiles available in $\mathcal{P}$, the fewer points may be tied at the top empirical quantile (leading to false positives). Two existing ranking methods are: 
\paragraph{1. Threshold Ranking}: The most common ranking approach, which we call Threshold ranking, identifies a dynamic outlier threshold per stream and returns a binary [0,1] classification based on values that exceed the threshold \cite{Lai2021, hundman2018detecting} that corresponds to a quantile in an empirical reference distribution $\mathcal{P}$ 

\paragraph{2. Sibling Ranking}: The other approach, which we call Sibling ranking, considers data from multiple, similar streams that share a parent, $g \in \mathcal{R}$ (e.g., $R_{sib}(g) = \{\text{parent}(r)=g | r \in \mathcal{R}\}$ and data considered is $d_{i,r}(t) \,\forall\, t \in [0, T], \,\forall\, r \in R_{sib}(g)$). Sibling ranking returns a real-valued, rankable score $\in [0, 1]$ that corresponds to the empirical quantile of new values compared to an empirical reference distribution $\mathcal{P}$ \cite{joshi2023computationally}\footnote{Sibling ranking reports mean scores for data in multiple sibling sets.}. 

Both Threshold and Sibling ranking return too many false positives because their $\mathcal{P}$ \textbf{1.} uses data from only one or a few streams that cannot readily be compared at a large scale, \textbf{2.} uses the whole stream's history instead of values that share temporal context, and \textbf{3.} only has a small and varying number of values in $\mathcal{P}$ due to limited data history per stream.

\section{Outshines Method}
\textbf{Overview:} First, Outshines (Fig. \ref{fig:Outshinesmet}) models hierarchical relationships in data streams for an indicator $i$. The resulting \textit{hierarchical network} facilitates outlier ranking by capturing contextual relationships across all data streams. For Delphi's data, this network contains streams from every region in the geospatial hierarchy, not just county-level streams. 

Then, Outshines takes as input the test statistics ($\phi$) output from any point univariate outlier detection method applied to these hierarchical streams for all relevant data points ($\phi(d) \,\forall\, d \in \{d_{i, r}(t) \,\forall\, r \in \mathcal{R}, t \in[t-\delta, t+\delta]\}$), where the nonstationarity and the available history of the data stream determine $\delta$. These resulting $\phi$ measure the degree of agreement between predicted and observed values, and extreme $\phi$ indicate a potential outlier. The tailored univariate method must ensure that the ranking of each $\phi$ matches that from $c$ \textit{per stream}. However, because some streams that are ill-suited to the tailored univariate method may consistently return extreme $\phi$, $\phi$ alone does not provide an informative ranking across many data streams and must be contextualized via a ranking algorithm.

Finally, Outshines outputs the rankable, scaled, real-valued quantiles of test statistics ($\phi$) from all regions in the hierarchy at time $t$ with respect to an empirical reference distribution generated per day, per indicator: $\mathcal{P}_{i, t}$. Outshines creates $\mathcal{P}_{i, t}$ by using hierarchical relationships and extreme value analysis on data from times close to $t$. 

Outshines addresses the sources of false positives present in  prior ranking methods because $\mathcal{P}_{i, t}$ \textbf{1.} contains data from streams all across the hierarchy and \textbf{2.} only considers data values at times similar to $t$. For Delphi's implementation of Outshines, where $t$ are days and the hierarchy is the geospatial hierarchy (Fig. \ref{fig:hiernetemp}), \textbf{3.} $\mathcal{P}_{i, t}$ contains more empirical quantiles than the other ranking algorithms and can thus provide higher resolution rankable scores.

\newcommand\NoDo{\renewcommand\algorithmicdo{}}
\begin{algorithm}
\caption{Outshines Ranking }
\label{alg}
Using Block Maxima to make $\mathcal{P}_{i,t}$ for indicator $i$ and day $t$\\
\textbf{Input}:$\phi(d_{i, r}(t)) \,\forall\, r \in \mathcal{R}$\\
\textbf{Output}:$y(d_{i, r}(t))\,\forall \,r\in \mathcal{R}$
\begin{algorithmic}[1]
\NoDo
\FOR{$\mathcal{R}_{sib} \in \mathcal{R}$:  \texttt{\#Stream Aggregation Dim.}} 
\STATE $P_{i, \mathcal{R}_{sib}, t} = \{\}$
\FOR{$h \in [t-14, t) \cup (t, t+14]$: \texttt{\#Temporal Dim.}}
\STATE \texttt{\#Block Maxima}
\STATE $P_{i, \mathcal{R}_{sib}, t}=P_{i,  \mathcal{R}_{sib}, t}\cup\max( \phi(d_{i, r}(h) | r\in \mathcal{R}_{sib}))$
\ENDFOR
\ENDFOR
\STATE $P_{i, t} = \cup_{\mathcal{R}_{sib} \in \mathcal{R}} P_{i,  \mathcal{R}_{sib}, t}$
\STATE $y(d_{i, r}(t)) \gets q( \phi(d_{i, r}(t)), P_{i, t})*\frac{\log(\lvert P_{i, t} \rvert)}{\log(\max{\lvert P_{i, t} \rvert})}  \,\forall\, r  \in \mathcal{R}$
\end{algorithmic}
\end{algorithm}

\subsection{Generating Outshines $P_{i, t}$} 
Outshines can rank outliers across multiple streams because of its empirical reference distribution, $P_{i, t}$. We generate this $P_{i,t}$ by adapting the block maxima technique from extreme value analysis. Traditional block maxima techniques split a data stream into equally-sized non-overlapping data blocks (e.g., one block per week) and calculate the maximum value in each block to form $\mathcal{P}$ \cite{gomes2015extreme} (e.g., $\mathcal{P}$ = $\{max(d_{i, r}(t$:$t$-7)),$max(d_{i, r}(t$-7:$t$-14$))...\}$). Highly-ranked outliers are points $d$ for which $\phi(d)$, the test statistic, is large even with respect to the reference distribution $\mathcal{P}$; i.e., they are more extreme than the most extreme value in a ``typical" week. The intuition for this approach is that if a univariate method is miscalibrated for a particular data stream and regularly returns $\phi$ with high values, $\mathcal{P}$ will contain many such examples, and outlier points must have even more extreme $\phi$ to stand out. However, traditional block maxima does not apply to streams with limited, nonstationary data \cite{rieder2014extreme}, like $\phi$ from Delphi's data. Even recent advances that address nonstationarity need long data histories \cite{zhang2021studying, sarmadi2022structural}. Our method addresses both limited data history and nonstationarity in streams of $\phi$ by changing both the block sizes and the aggregation strategy for the block maxima technique. 

\paragraph{Creating New Blocks (Alg 1. lines 1-7):} Each block in block-maxima has a temporal dimension (e.g., day, week, month) and a stream aggregation dimension (typically one stream). Aggregating homogenous, or similar, streams per block is a known way to calculate block maxima over more data, but identifying an appropriate homogeneity test is difficult and could result in sets of dissimilar streams \cite{lilienthal2022note}. Instead, we identify homogenous streams as those that share a parent $ r\in \mathcal{R}_{sib}(g)$. Each set of sibling streams, $R_{sib}$, in $\mathcal{R}$ is subject to similar conditions by being in the same tier and having the same parent\footnote{HRRs at state borders belong to multiple sibling sets, which models that these regions are subject to potential outlier events from either of their parent streams.}. Aggregating streams across $\mathcal{R}_{sib}$ creates blocks of similar regions and addresses limited history. 

We then use two mechanisms to address nonstationarity. First, our blocks have a temporal dimension of one observation. Limiting the temporal dimension like this creates blocks across streams instead of accross time so that temporal variations (e.g., weekday effects) do not skew $\mathcal{P}_{i, t}$, but there are still enough observations for block maxima (e.g., our blocks are {7 sibling streams x 1 day} vs. the traditional {1 stream x 7 days}). Second, we limit the temporal range of blocks to data similar to the time being evaluated (a regime). Only using data from the same regime ($t \pm \delta$) ensures $P_{i, t}$ is relevant to data at time t. For Delphi, that regime is 28 days, with 14 days of historical data and 14 days of prospective days (when data after t is available), so these blocks are the most similar to the time considered. This 28-day range (4 weeks) is a standard followed by many organizations participating in the CDC's respiratory illness forecasting tasks \cite{biggerstaff2018results}. 

We can then continue with the standard block maxima procedure on these modified blocks so that $\mathcal{P}_{i, \mathcal{R}_{sib}, t}$ contains the maximum $\phi$ per indicator, per $R_{sib}$ per $t$ (Alg 1. line 5). We use the maxima (the most common choice in extreme value analysis) because we want values that describe the empirical distribution at the tail, and intermediate quantiles are ill-defined for small sibling groups (e.g., 95\% from a set of 3 regions). 

\paragraph{Aggregating Regions and Scaling (Alg 1. lines 8-9):} To make an empirical reference distribution with many observations that capture extreme $\phi$ from all $\mathcal{R}_{sib}$, these $\mathcal{P}_{i, \mathcal{R}_{sib}, t}$ for $\mathcal{R}_{sib} \in \mathcal{R}$ can be pooled together to create $\mathcal{P}_{i, t}$. Thus, $\mathcal{P}_{i, t}$ represents the distribution of recent extreme $\phi$ equally weighted from each set of geospatial regions, as shown in Alg. \ref{alg}, line 8. Finally, Outshines outputs a score using the quantile ($q$) of new $\phi$ at time $t$ of $\mathcal{P}_{i, t}$. 

We apply Outshines to each of Delphi's indicators separately so that computation can occur across all indicators (in parallel) and so that there are no assumptions about the relationships between the constantly changing set of available indicators. However, scores from $\mathcal{P}_{i, t}$ across different indicators with more observations should be weighted higher because they have a higher resolution. Thus, Outshines scales each quantile by $\lvert \mathcal{P}_{i, t} \rvert$ divided by the log of the maximum possible observations, (e.g., $\log(|\mathcal{R}_{sib} \in \mathcal{R}|*$regime)) to return a score ($y$) in [0, 1]. The log factor shrinks the scale for both the numerator (points available) and denominator (total possible points) so that between two indicators with different possible total lengths (denominator) with the same available/total points, the one with the longer stream will have the higher multiplicative factor (and might be ranked higher). 

\begin{table*}[ht]
\begin{threeparttable}
{\begin{tabular}[b]
{rlrrrrrr}
\hline
\multicolumn{3}{c}{\multirow{2}{*}{\textbf{\textit{Task}}}}
 & \multicolumn{4}{c}{\textbf{\textit{Ranking Method}}}\\
  &&&\makecell[c]{Thresh.}& \makecell[c]{Opt. Thresh.} & \makecell[c]{Sibling} & \makecell[c]{Outshines} \\
\midrule
\multicolumn{2}{l}{\textbf{Timing/Indicator (s)}} &\makecell[l]{{Generate $\phi$}}&&&&\\
\multirow{6}{*}{\rotatebox[origin=c]{90}{\parbox[c]{1.1cm}{\centering \textbf{\textit{UOD}}}}}& \cellcolor{yellow!25}\makecell[l]{EWMA} &  \cellcolor{yellow!25}57.9 $\pm$ 35.17 &\cellcolor{yellow!25}* 6.71 $\pm$ 2.59 &\cellcolor{yellow!25}* 6.6 $\pm$ 2.51 &\cellcolor{yellow!25} 319.67 $\pm$ 172.55 &\cellcolor{yellow!25} 50.58 $\pm$ 46.40 \\
&\makecell[l]{FlaSH} & 458.81 $\pm$ 146.21 &* 5.33 $\pm$ 2.32& * 5.2 $\pm$ 2.12& 326.54 $\pm$ 160.64 & 45.85 $\pm$ 46.84 \\
&\makecell[l]{AR }& 36.13 $\pm$ 19.85 & 5.61 $\pm$ 3.92 & 4.32 $\pm$ 3.02 & 270.61 $\pm$ 156.14 & 58.97 $\pm$ 47.10 \\
&\makecell[l]{Isolation Forest}& 420.59 $\pm$ 270.15 & 65.04 $\pm$ 43.16 & 61.84 $\pm$ 41.92 & 190.58 $\pm$ 104.08 & 39.06 $\pm$ 29.10 \\
&\makecell[l]{DeepLog}& 6520 $\pm$ 4398 & 53.2 $\pm$ 36.13 & 52.79 $\pm$ 35.95 & 188.78 $\pm$ 102.44 & 38.97 $\pm$ 28.97 \\
&\makecell[l]{Telemanom} & 6160 $\pm$ 4449 & 64.65 $\pm$ 47.08 & 68.26 $\pm$ 52.38 & 293.36 $\pm$ 159.5 & 57.75 $\pm$ 43.44 \\

\multicolumn{2}{l}{\multirow{1}{*}{\textbf{$\#$ Ties/Indicators}}} &&&&&\\

\multirow{6}{*}{\rotatebox[origin=c]{90}{\parbox[c]{1.1cm}{\centering \textbf{\textit{UOD}}}}} &\makecell[l]{\cellcolor{yellow!25}EWMA} &\makecell[c]-&\cellcolor{yellow!25}*15.81\textbf{k} $\pm$ 1.97\textbf{k} &\cellcolor{yellow!25}* 6.05\textbf{k} $\pm$ 2.54\textbf{k} &\cellcolor{yellow!25} 585.33 $\pm$ 549.13 &\cellcolor{yellow!25} \cellcolor{yellow!25} 6.67 $\pm$ 0.65\\
&\makecell[l]{FlaSH} &\makecell[c]-& * 22.02\textbf{k} $\pm$ 1.97\textbf{k}& * 8.08\textbf{k}$\pm$ 2.13\textbf{k}& 159.33 $\pm$ 23.37 & 7.67 $\pm$ 2.85\\
&\makecell[l]{AR}&\makecell[c]-& 42.11\textbf{k} $\pm$ 18.84\textbf{k} & 7.02k $\pm$ 3.714k & 127.67 $\pm$ 18.29 & 11.67 $\pm$ 10.51 \\
&\makecell[l]{Isolation Forest}&\makecell[c]-& 89.87\textbf{k} $\pm$ 67.23\textbf{k} & 39.80\textbf{k} $\pm$ 21.93\textbf{k} & 3.87\textbf{k} $\pm$ 2.23\textbf{k} & 20.67 $\pm$ 20.88 \\
&\makecell[l]{DeepLog}&\makecell[c]-& 32.29\textbf{k} $\pm$ 26.07\textbf{k} & 14.24\textbf{k} $\pm$ 3.44\textbf{k} & 260.33 $\pm$ 63.78 & 18.0 $\pm$ 2.99 \\
&\makecell[l]{Telemanom}&\makecell[c]-& 78.78\textbf{k} $\pm$ 39.10\textbf{k} & 53.31\textbf{k} $\pm$ 32.42\textbf{k} & 215.0 $\pm$ 89.44 & 14.0 $\pm$ 2.26 \\
\bottomrule
\end{tabular}}
\caption{Baseline Comparisons - Yellow highlights the deployed combination. This combination has the fewest of ties.}
\label{table:eval}
\end{threeparttable}
\end{table*}

\paragraph{Ranking Algorithm Comparison:} The design of Outshines addresses the sources of false positives. First, Outshines ensures that $\mathcal{P}_{i,t}$ does not over-represent any region or time so that the output scores are comparable. Outshines compares every $\phi(d_{i, r}(t)) \,\forall\, r \in \mathcal{R}$ to the same $\mathcal{P}_{i, t}$, unlike Sibling or Threshold ranking, where the number of observations in $\mathcal{P}$ varies depending on missing data per stream. Outshines also only uses data within a regime to create $\mathcal{P}_{i, t}$. Finally, for Delphi's data, Outshines has more granular output scores because it has more observations that characterize $\mathcal{P}$, as follows: 

\begin{enumerate}
    \item Threshold ranking contains all $\phi$ from a single stream so $(\mathcal{P}_{i, r}:= \{\phi(d_{i, r}(t)) | t \in [0, t) \cup (t, T] \})$. 
    \item Sibling ranking $\mathcal{P}$ is at least as large because it contains all $\phi$ from sibling streams $(\mathcal{P}_{i, \mathcal{R}_{sib}}$:=$\{\phi(d_{i, r}(t)) | r\in{\mathcal{R}_{sib}}, t \in [0, t) \cup (t, T]\})$
    \item Outshines ranking $\mathcal{P}$ is $\lvert \mathcal{P}_{i, t}\lvert  = \lvert R_{sib} \in \mathcal{R}\rvert \times 28$
\end{enumerate}
On Delphi's data, HRR and HHS tiers create intuitive and numerous $R_{sib}$ (369 $R_{sib}$ with an average size of 15.85 regions) and $T \leq 300$. Thus, Outshines $\mathcal{P}$ is twice as large as sibling ranking
while only using a regime of 28 days, as previously described. Increasing the regime from 28 can add more observations and increase the granularity of Outshines' output scores. However, this could lead to false positives as it reduces $\mathcal{P}_{i,t}$'s temporal similarity to $t$. Accordingly, before the evaluation began, we preregistered the Github commit of our Outshines implementation with these parameters on {OSF} \cite{joshi2023large}. 

\section{Tailored Univariate Method (EWMA)}
Outshines can rank $\phi$ produced by any univariate outlier detection method that matches the expert relative ranking per stream. We provide one such univariate method tailored to Delphi's experts in Summer 2023 (as shown in Fig. \ref{fig:Outshinesmet} Box 2). At that time, based on our interviews and observations, these experts highly prioritized phenomena that manifested as outliers representing rapid changes in streams with a high population, few historical outliers, and few missing values (high $c$). We tailored a model-free Exponentially Weighted Moving Average (EWMA) method to identify such points in the public health setting \cite{burkom2020essence}).

\textbf{Prediction Generation:}  We use a kernel weighting function, $\mathcal{K}$(t) whose value in position $w$ is 
\[
\mathcal{K}_w(t) =  
\begin{cases}
  0 & \text{ if } w = t\\
  e^{-\frac{|w-t|}{\tau}} & \text{otherwise}
\end{cases}
\]
where $\tau$=2 to prioritize the temporally closest data values. $\mathcal{K}$  is discretely convolved with the stream values $d_{i,r}(t) \,\forall\, t \in [0, T]$ for stream $i,r$ and then standardized to generate stream predictions $\hat{d}_{i,r}(t)$: 

\[\hat{d}_{i,r}(t)= (\mathcal{K}*d_{i,r})(t)/\sum{\mathcal{K}(t)}
 \]

\textbf{Test Statistic ($\phi$) Calculation:} The test statistic ($\phi$) compares observed and predicted values, as described following this paragraph. To cacluate $\phi$, we define and scale $l(t)$=$\hat{d}_{i,r}(t) - d_{i,r}(t)$) to account for stream outlier history. Taking the absolute value ensures block maxima selects points representing rapid change regardless of direction. Lastly, we multiply this value by the available stream history and population ($r_{pop}$) because these features are important to experts; an outlier in $r_{pop}$= 100k should outrank a national outlier if it is at least 2x as important\footnote{In smaller $r_{pop}$, using absolute difference for $\phi$'s does not capture subtle outliers (e.g. going from 4 to 5 cases is more concerning than 34 to 35), but can be used to detect the extreme outliers Delphi experts cared about in July 2023.}.

\[\phi_{i,r}(t) = |\frac{l_{i,r}(t)-\small\text{median}({l_{i,r}}(t))}{\sigma_{l_{i,r}}}|\cdot\log (|d_{i, r}|)\cdot\log(r_{pop})\] 

During deployment, we observed that EWMA's $\phi$ for inflection points and points in trends longer than a week did not match that from $c$. However, Delphi's experts preferred EWMA to a more complex but more accurate method because EWMA only requires a linear pass over each stream in parallel.

\begin{figure*}
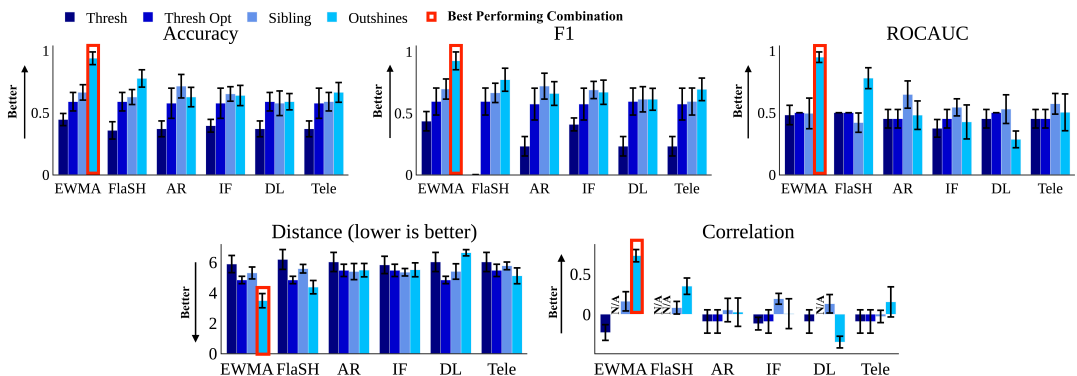

\centering
\includegraphics[width=0.8\textwidth]{metrics_binary-final.pdf}\hfil
\includegraphics[width=0.55\textwidth]{metrics_ranking-final.pdf}\hfil
\caption{Evaluation Metrics: The red box highlights the best performing combination ranking method x tailored univariate method for standard binary and ranking metrics. Correlations were N/A when the method returned all 0s or 1s.}
\label{fig:metric}
\end{figure*}

\section{Evaluations}

To measure Outshines' correctness, we used standard outlier detection metrics comparing the expert-labeled data we collected on real public health data streams to different combinations of ranking x univariate methods (Evaluation 1). We also monitored Outshines' deployment over 4 months and reported results from July to August 2023 (Evaluation 2).

\subsection{Evaluation 1. Comparing Method Combinations}
We asked Delphi's experts to rank outliers from real data streams analyzed with Outshines during deployment on July 27, 2023 (snapshot dataset). This data set contained 6383 streams from the top three indicators which had outliers worth investigating in June 2023\footnote{1. Outpatient doctor visits for COVID-related symptoms, 2. \% COVID-positive antigen tests, and 3. Estimated \% of new COVID hospital admissions based on claims data} (and were thus likely to contain rankable outliers during the snapshot). To compare EWMA with model-based methods, we provided 200 days of training data and computed rankings over the last 100 days for a total of 1.4 million data points over 300 days.

We evaluated the performance of all combinations of ranking x univariate methods on this snapshot dataset compared to the expert-labeled data we collected. The four implemented ranking methods were: 
\begin{itemize}
\item Threshold ranking \cite{Lai2021}
\item Optimized Threshold ranking\footnote{The TODS implementation threshold=0.9, but we set it to 0.99 to match the frequency outliers are expected \cite{wu2021current}}
\item Sibling ranking \cite{joshi2023computationally}
\item Outshines ranking (see Methods)
\end{itemize}
Comparing these ranking methods served as an ablation for grouping sibling streams (Threshold to Sibling) and using extreme values across hierarchies (Sibling to Outshines). 

The univariate methods (UOD) were the Tailored Univariate Method EWMA (see Methods), FlaSH \cite{joshi2023computationally}, an outlier detection algorithm designed for smaller-scale public health streams, and the TODS implementations \cite{Lai2021} of Telemanom \cite{hundman2018detecting} (Tele) \& DeepLog \cite{du2017deeplog} (DL) (State of the art deep learning methods), Isolation Forest (IF): \cite{liu2008isolation}, and Linear AutoRegressive Models (AR) \cite{gupta2013outlier}. Our proposed combination is \textbf{Outshines x EWMA}, and our experimental design (e.g., analysis plan, web forms, implementations) was preregistered before data collection began. These experiments were conducted on a 2.6GHz Intel Core i7 machine with OSX in Python3.8 for TODS compatibility. 

\paragraph{Combination Feasibility} The first section of Table \ref{table:eval} : Timing/Indicator (s)\footnote{Because EWMA and FlaSH are not in TODS, authors implemented a similar ranking method for comparison (indicated by *).} shows that Outshines is more than 4.5x faster than Sibling ranking and generally faster than TODS Threshold ranking. Because Outshines computes $P_{i,t}$ across a 28-day window of $\phi$ (parallel)  instead of all historical $\phi$ in a stream (serial), there may be more pronounced performance gains as $T$ increases. Further, our chosen combination of Outshines x EWMA took much less time than deep learning univariate methods, which would have required 4x Delphi's compute resources to generate $\phi$ on deployment data (updated daily) in July 2023 (55 indicators vs. 3 in the snapshot set) to keep daily processing times under one day. 

More importantly, the second part of Table \ref{table:eval} : $\#$ Ties/Indicator shows that Outshines is also the \textit{only} ranking method that produced few enough maximally tied outliers for reviewers to investigate daily across all the univariate method combinations. Outshines reduces Sibling ranking's maximally tied outliers by over 10x (typically 6-18 ties vs. 200-3k ties). If Outshines' rankings are meaningful, these results show that Outshines can promptly prioritize outliers from large-scale public health streams. 

\subsubsection{Expert Ranking Generation and Comparison}
To identify if Outshines ranks were prioritizing the outliers experts wanted, we needed to address common limitations in obtaining expert-labeled outlier rankings \cite{wu2021current} by having a large number (n = 17 vs. n=3 \cite{wong2004data}) of experts, who regularly use public health data streams for research or development purposes, evaluate outliers from real data streams (and not synthetic data \cite{lai2021revisiting}). These experts could only review a few of the millions of data points in the snapshot dataset, and a random sample of that size would likely contain no outliers worth ranking. Instead, because the top deployed Outshines x EWMA rankings are empirically a subset of points of the deployed Sibling x EWMA rankings, we could directly compare the two methods by asking experts to rank points that were highly differentiated by Outshines in deployment (one example per decile, as available). We designed a survey where experts inspected two interactive web pages, each with 5 data streams, corresponding to the selected outliers and ranked them using the scale \{5: Most important, 1: Least important\}. We had one control stream in each web page to measure internal consistency. Using the ranky package \cite{pavao2020ranky}, we found that respondents demonstrated internal consistency with high Spearman correlation ($0.85 \pm 0.27$) and low Euclidian distance ($1.21 \pm 0.13$) in pairwise rankings for the control stream. Reviewers also demonstrated reasonable interrater consistency, with a high Spearman correlation (0.71) for rankings \textit{across} respondents.

In Fig. \ref{fig:metric}, we display mean metrics per stream per person with a 95 \% CI error bar for each combination of ranking x univariate method. We calculated binary metrics (Accuracy, F1, and ROCAUC) using the top-k points as the positive class, where k is the number of streams with a ranked outlier per person, and ranking metrics (Swap Correlation (higher is better) and Hamming Distance (lower is better)) by using the respondent's absolute ranking. 

Our results show that, of all combinations, Outshines x EWMA scores best match the experts for these standard metrics. For example, the mean AUC of the Outshines x EWMA combination is 0.95 vs. 0.78 from the next best combination (Outshines x FlaSH). Similarly, Outshines x EWMA has the highest mean correlation at 0.73 vs. the next highest from any other ranking method (Sibling x DL) at 0.13. Outshines also performed well with other univariate methods -  Outshines x FlaSH is always the second best performing combination. After Outshines, the second best performing ranking method is Sibling ranking, which typically has the third-best performance (usually Sibling x EWMA). 

\subsection{Evaluation 2: Outshines' Deployed Performance}
Experts in Delphi have used Outshines in their daily outlier review process since April 2023. First, they receive the Outshines scores $y(d)$ for all recently updated data. Second, they review the top 20-25 highest ranking points and record any outliers worth investigating. Then, they investigate these points by analyzing patterns in the national, parent, sibling, or child streams. Afterward, they meet with the research team weekly to discuss any findings. 

\textbf{A. Quantitative Results:} Between July 10-August 5, the daily deployment data volume from 55 currently updated indicators was 3.5 million $\pm$ 280k points. Applying Outshines x EWMA over this data volume took $123.44 \pm 189.16$ minutes per day on production hardware (3.3GHz Intel Core i7 machine with Pop!OS, Python3.10), and the number of ties generated over this data using Outshines ranking ($21 \pm 5.5$) were much fewer than with Sibling ranking ($14\textbf{k} \pm 1.7\textbf{k}$).

\begin{figure}
  \centering
  \includegraphics[width=7.5cm]{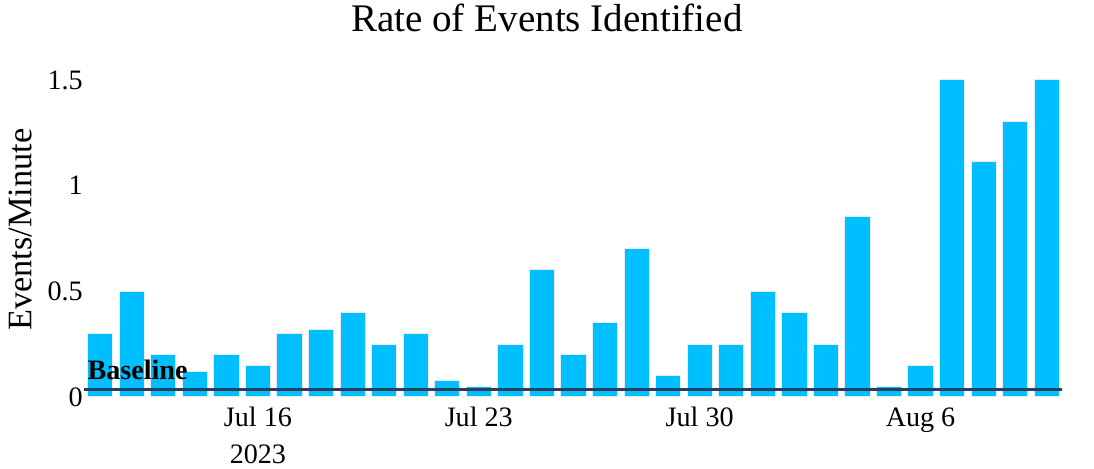}\hfil
  \caption{Experts can identify outliers worth investigating more quickly with Outshines.}
  \label{fig:loads}
\end{figure}

The most important metric for Delphi was how quickly experts identified outliers worth investigating (events) compared to a baseline from volunteers who met weekly to manually review the data without an outlier detection method. As per their meeting notes, volunteers found 0-1 outliers worth investigating in this baseline every 30 minutes (0.033 Events/Minute). Fig. \ref{fig:loads} shows that our combination of Outshines x EWMA increased the rate of expert event identification ($0.43 \pm 0.14$ events, which is an improvement of $9.13 \pm 2.26$\textbf{x} over the baseline), as determined by self-reported review time and the number of outliers marked for investigation. Further, the average number of outliers marked daily with Outshines, 6, far exceeds baseline efforts of 0-1 outliers during manual review. 

Additionally, for one week, experts first reviewed 10 random maximally tied points from Sibling ranking and then the top 10 data points from Outshines ranking for a more direct comparison between these ranking methods. In this evaluation, experts found 4.02\textbf{x} as many outliers using Outshines than Sibling ranking and at $3.96 \pm 1.27$\textbf{x} the rate. 

\textbf{B. Qualitative Results:} Outshines provided Delphi's first ever insight into outlier data points worth investigating on a large scale. Experts preferred conducting a manual review to analyzing a random sample of thousands of outliers, like those produced by Sibling or Threshold ranking. This means that Outshines was the only ranking method that Delphi could deploy in practice! Experts also felt that the feedback cycle with the research team was crucial in staving algorithm aversion \cite{dietvorst2015algorithm}, which occurs when humans no longer trust an algorithm output (and would prefer manual review). These iterations ensured that the tailored EWMA combination matched expert needs (which Outshines scaled) and will continue to be useful as Delphi curates new data.  

\section{Conclusion}
Informative outlier rankings enable disease control experts to analyze large-scale public health streams. We modeled this problem as a new task of ranking test statistics output from tailored univariate outlier detection methods (multi-stream outlier ranking). Our method, Outshines, uses hierarchical relationships in data streams and modified extreme value analysis for this task. Our expert evaluations and performance experiments show that Outshines outperforms other ranking methods on traditional metrics. Its deployed performance also met experts' needs by enabling them to identify outliers worth investigating 9.1x faster than the baseline. As public health data volumes grow, Outshines provides experts with a scalable, flexible, and accurate way to find the most important outliers quickly.

\paragraph{Acknowledgements:} This work was supported by a cooperative agreement funded solely by CDC/HHS under federal award identification number U01IP001121, “Delphi Influenza Forecasting Center of Excellence” and by the  by the National Science Foundation Graduate Research Fellowship under Grant No. DGE1745016 and DGE2140739. The contents do not necessarily represent the official views of, nor an endorsement by, CDC/HHS/NSF or the U.S. Government. 

\bibliographystyle{acm}
\bibliography{sources}
\end{document}